\documentclass[letterpaper]{article} 
\usepackage[preprint]{aaai2027}

\usepackage[hyphens]{url} 
\usepackage{graphicx} 
\usepackage{natbib} 
\usepackage{caption} 
\usepackage{booktabs}
\usepackage{amsmath}
\usepackage{amssymb}

\title{PMMC: Prospective Multimodal Memory Compilation for Long-Term LVLM Agents}

\author{
    Jingyu Sun\textsuperscript{\rm 1,\rm 2},
    Yan Lin\textsuperscript{\rm 3},
    Yuyang Xue\textsuperscript{\rm 4},
    Yifan Wang\textsuperscript{\rm 1},
    Zhengtao Yao\textsuperscript{\rm 5},\\
    Rui Qian\textsuperscript{\rm 6},
    Zefeng Xu\textsuperscript{\rm 1},
    Jiachen Li\textsuperscript{\rm 7},
    Xianyang Liu\textsuperscript{\rm 8},
    Jiancheng Pan\textsuperscript{\rm 9},\\
    Jingyuan Sun\textsuperscript{\rm 1},
    Syed Murtuza Baker\textsuperscript{\rm 1},
    Hongpeng Zhou\textsuperscript{\rm 1}\corresponding
}
\affiliations{
    \textsuperscript{\rm 1}The University of Manchester \quad
    \textsuperscript{\rm 2}The University of Melbourne \quad
    \textsuperscript{\rm 3}The University of Newcastle\\
    \textsuperscript{\rm 4}The University of Edinburgh \quad
    \textsuperscript{\rm 5}The University of Southern California \quad
    \textsuperscript{\rm 6}Fudan University\\
    \textsuperscript{\rm 7}The University of Texas at Austin \quad
    \textsuperscript{\rm 8}Independent Researcher \quad
    \textsuperscript{\rm 9}Ant Group
}

\begin{document}

\maketitle

\begin{abstract}
Long-term memory is essential for LVLM agents to maintain consistency and integrate information across extended multimodal interactions. Existing agent memory systems, however, often reduce visual experiences into textual summaries or rely on static retrieve-then-reason pipelines, which are inefficient at query time and brittle when questions require image-text binding, temporal updates, or visual details. We propose Prospective Multimodal Memory Compilation, a framework that shifts part of the memory reasoning process from query time to memory consolidation time. Given accumulated multimodal interactions, a Questioner predicts future question candidates, a Planner compiles question-conditioned multimodal memory programs, and a Doubter verifies whether the planned evidence path can support the predicted answer. The verified question-program pairs form a structured question bank for efficient query-time routing and evidence retrieval. Experiments on multimodal long-term memory benchmarks show that our method improves answer quality and visual evidence recall while reducing query-time token and latency costs. Extensive ablations analyze the effects of self-feedback, dynamic planning, raw-image access, and question bank coverage. 
\end{abstract}

\section{Introduction}

Long-term memory is essential for large vision-language model (LVLM)
agents operating over extended multimodal interactions. Unlike
conventional visual question answering, agent experiences arrive
incrementally across sessions, so evidence for a future query may be
distributed across distant turns, images, and later corrections. A
capable agent must therefore recall earlier information, bind visual
entities to linguistic references, combine evidence across events, and
update stale beliefs. Recent benchmarks show that current systems
remain challenged by long-range multimodal reasoning, temporal updates,
conflicts, and unsupported queries
\cite{maharana2024evaluating,ren2026memlens,bei2026mem}.

Existing approaches broadly rely on either long-context LVLMs or
external memory. Long-context models directly consume interleaved
histories, but their computational cost grows with context length and
their ability to locate relevant evidence often degrades as histories
expand
\cite{song2024milebench,wang2026mmlongbench,ren2026memlens}.
Memory-augmented agents instead store and selectively retrieve past
experiences using summaries, structured notes, hierarchical stores, or
graphs
\cite{lewis2020retrieval,park2023generative,packer2023memgpt,
xu2026mem,chhikara2025mem0}.
Although recent multimodal systems preserve richer visual and
personalized information
\cite{long2025seeing,feng2026m2a}, they still commit to a memory
representation and access policy before future queries are known.

\begin{figure*}[t]
    \centering
    \includegraphics[width=0.95\textwidth]{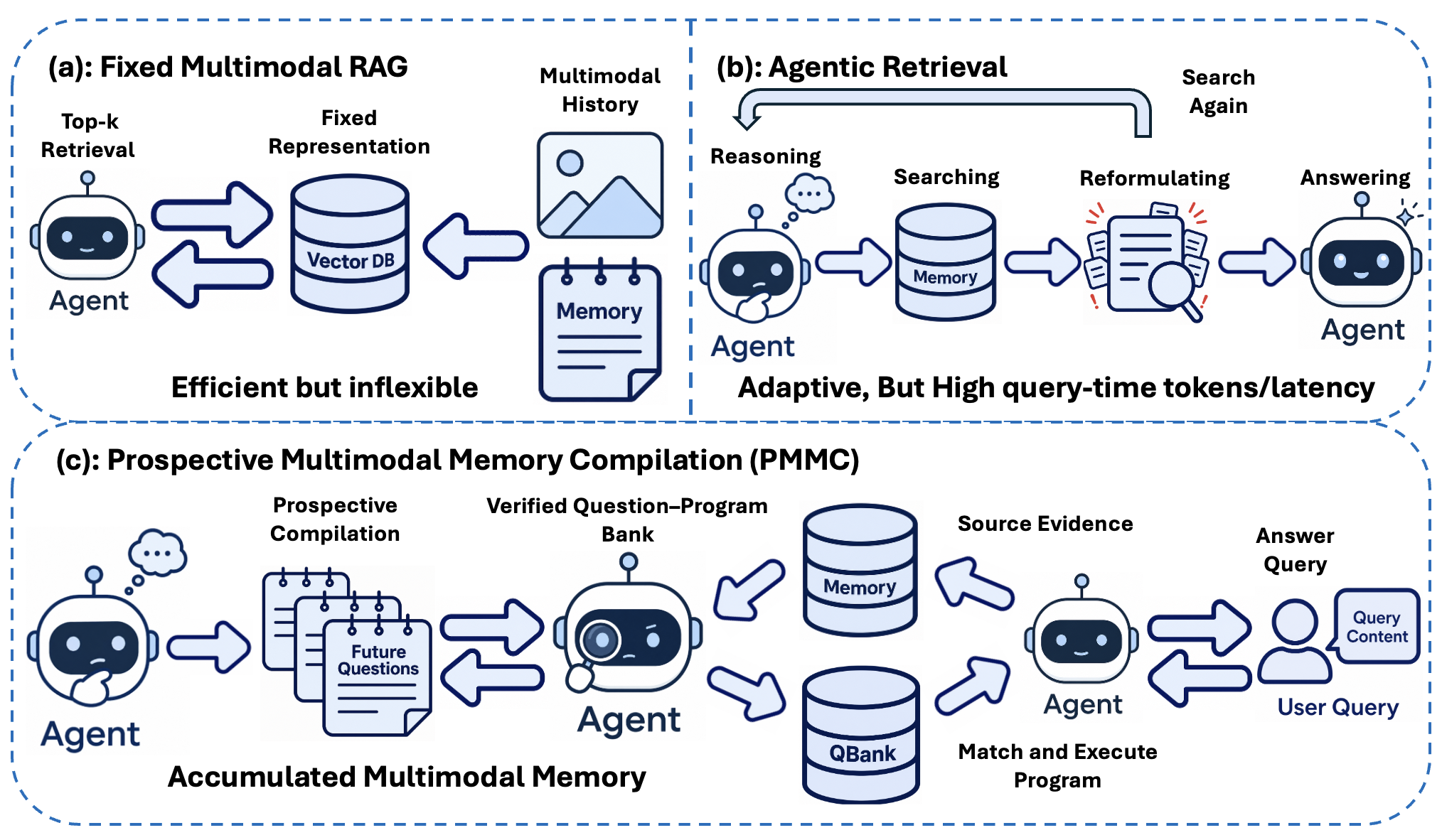}
    \caption{
    \textbf{Comparison of multimodal memory-access paradigms.} Fixed RAG uses a predefined memory representation and retrieval policy, whereas
    agentic retrieval adapts through repeated query-time reasoning.
    PMMC instead compiles and verifies question-conditioned memory
    programs during consolidation and executes them online, retaining
    multimodal RAG as a fallback.
    }
    \label{fig:motivation}
\end{figure*}

This commitment is especially problematic for multimodal memory.
Caption-based stores are efficient but permanently lose visual
attributes omitted during captioning. Contrastive encoders such as
CLIP and SigLIP align images and text semantically
\cite{radford2021learning,zhai2023sigmoid}, yet do not explicitly
preserve interaction-specific bindings, such as a subsequently assigned
name for an earlier image. Unified multimodal encoders can represent
mixed inputs
\cite{li2026qwen3}, but still require a fixed granularity for
compressing heterogeneous evidence. A single representation is
unlikely to suit all information needs: factual questions may require
text, entity questions image--text associations, visual-detail
questions original pixels, and temporal or multi-hop questions several
linked memories. Existing benchmarks accordingly report losses from
caption-only compression and from retrieving redundant or weakly
aligned multimodal evidence
\cite{bei2026mem,ren2026memlens}.

Conventional RAG resolves these choices only after a query arrives by
retrieving, optionally reranking, and supplying a fixed number of
candidates
\cite{lewis2020retrieval}. Agentic retrieval is more adaptive, allowing
models to reformulate searches or interleave retrieval with reasoning
\cite{yao2022react,trivedi2023interleaving,jiang2023active,
asai2024self,jeong2024adaptive}, but requires additional online model
calls and retrieval rounds. Index-time question generation can improve
alignment between text documents and likely queries
\cite{lewis2021paq,neeser2025quote}, yet existing approaches primarily
use generated questions as retrieval keys rather than compiling and
verifying executable multimodal access procedures.

We therefore ask whether an agent can anticipate plausible future
information needs during memory consolidation and prepare how their
evidence should be accessed. We propose
\textbf{Prospective Multimodal Memory Compilation (PMMC)}, illustrated
in Figure~\ref{fig:motivation}. A \emph{Questioner} generates
source-grounded prospective questions, a \emph{Planner} compiles each
into a typed multimodal memory program, and a \emph{Doubter} verifies
the program through execution and bounded iterative refinement.
Accepted question--program pairs form a structured Question Bank. At
query time, PMMC matches the incoming query, executes the selected
frozen access strategy over the currently visible memory, and answers
from the recovered source evidence without invoking the compilation
agents online. Uncertain routes or insufficient evidence instead invoke
multimodal RAG. The Question Bank is therefore a routing and program
index, not an answer cache.

Across MEMLENS and Mem-Gallery with four LVLM backbones, PMMC ranks
first in six of eight configurations and improves the overall
macro-average Harmonized Judge score by 3.2 points over the strongest
baseline (Table~\ref{tab:main_results}). Our contributions are
threefold:
\begin{itemize}
    \item We formulate \emph{prospective multimodal memory compilation},
    moving question-conditioned access planning from query time to
    memory-consolidation time while preserving source-grounded
    answering.

    \item We introduce a Questioner--Planner--Doubter framework that
    compiles and execution-validates typed multimodal memory programs,
    with online fallback when prospective coverage is insufficient.

    \item We evaluate PMMC against long-context, fixed-retrieval, and
    multimodal agent-memory baselines, and analyze self-feedback,
    dynamic planning, raw-image access, and write--query cost
    trade-offs.
\end{itemize}

\section{Methodology}
\label{sec:methodology}

Prospective Multimodal Memory Compilation (PMMC) shifts part of memory
access planning from query time to memory consolidation time. Given the
multimodal history accumulated so far, PMMC generates plausible future
information needs, compiles each into an executable memory program, and
verifies the program through actual execution. The accepted
question--program pairs form a runtime Question Bank. When a real query
arrives, PMMC retrieves and executes a compiled program over the
currently visible memory; uncertain routes or insufficient evidence
invoke multimodal RAG fallback~\cite{lewis2020retrieval}.

\begin{figure*}[t]
    \centering
    \includegraphics[width=0.95\textwidth]{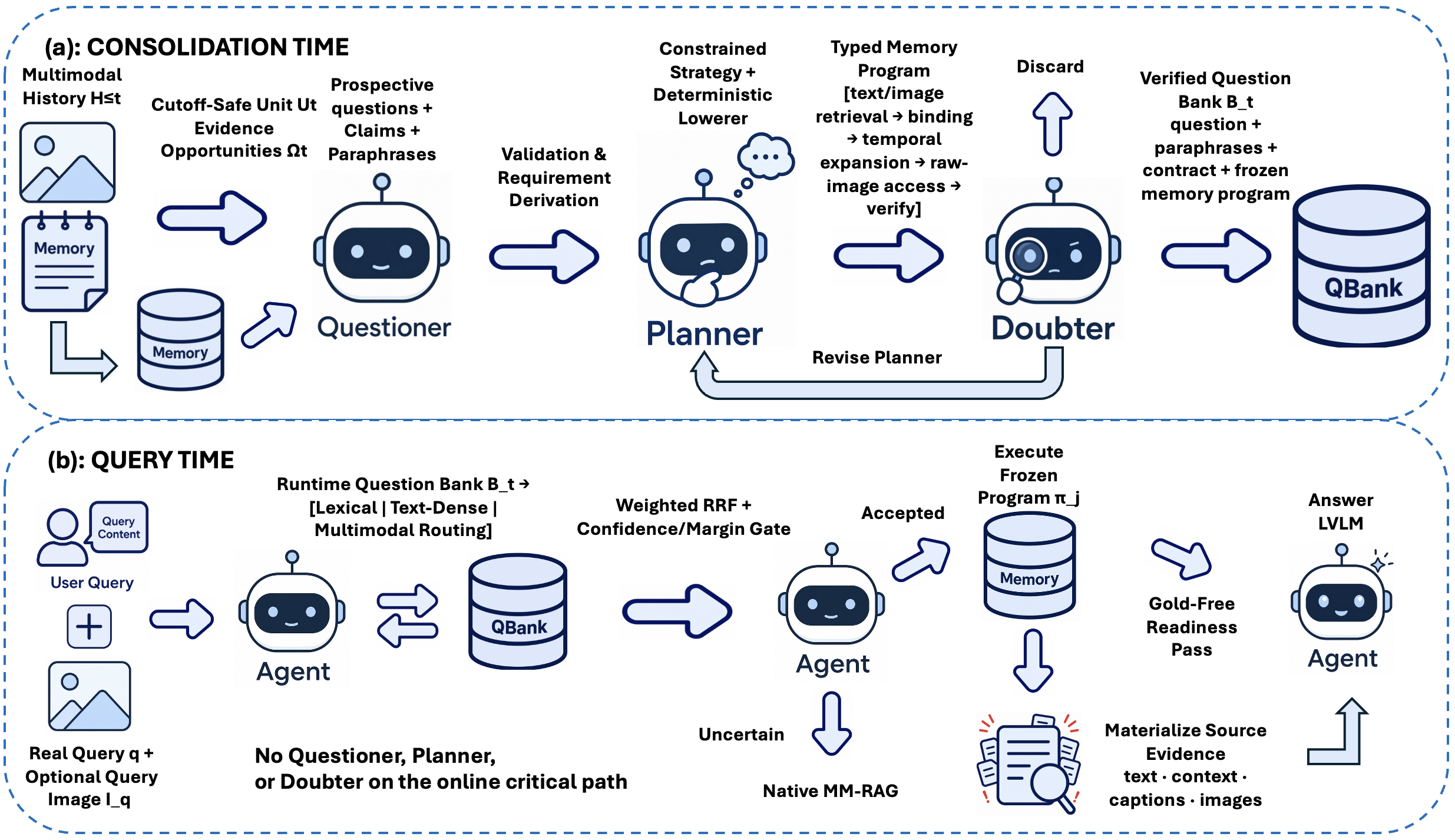}
    \caption{
    \textbf{Overview of PMMC.} During consolidation, a Questioner generates
    source-grounded prospective questions, a Planner compiles typed
    memory programs, and a Doubter verifies them through execution and
    bounded revision. Compiler-private provisional answers and source
    support are removed before accepted question--program pairs enter
    the runtime Question Bank. At query time, PMMC routes the incoming
    query to a frozen program, re-executes it over the currently visible
    memory, and answers from the recovered source evidence. Failed
    routing or readiness checks invoke multimodal RAG fallback; no
    Questioner, Planner, or Doubter call occurs online.
    }
    \label{fig:pmmc_pipeline}
\end{figure*}

\begin{table*}[t]
    \centering
    {%
    \small
    \setlength{\tabcolsep}{2.4pt}
    \renewcommand{\arraystretch}{1.06}

    \begin{tabular}{lccccccccccc}
        \toprule
        Method
        & \multicolumn{2}{c}{GPT-5-mini}
        & \multicolumn{2}{c}{Claude Haiku 4.5}
        & \multicolumn{2}{c}{Qwen3.5-9B}
        & \multicolumn{2}{c}{Qwen3.5-27B}
        & \multicolumn{3}{c}{Macro Average} \\
        \cmidrule(lr){2-3}
        \cmidrule(lr){4-5}
        \cmidrule(lr){6-7}
        \cmidrule(lr){8-9}
        \cmidrule(lr){10-12}
        & ML & MG
        & ML & MG
        & ML & MG
        & ML & MG
        & ML Avg. & MG Avg. & Overall \\
        \midrule

        Full-Ctx-Caption
            & 39.7
            & 37.5
            & 35.6
            & 37.2
            & 47.8
            & 36.4
            & \underline{51.4}
            & 41.2
            & 43.6
            & 38.1
            & 40.9 \\

        Full-Ctx-Raw
            & 36.3
            & 33.5
            & 32.7
            & 34.5
            & 46.4
            & 32.5
            & 47.7
            & 39.6
            & 40.8
            & 35.0
            & 37.9 \\

        \addlinespace[0.5mm]

        Caption-Text RAG
            & \textbf{45.6}
            & \underline{47.4}
            & \underline{39.8}
            & 36.6
            & \textbf{58.4}
            & \underline{53.2}
            & 48.7
            & 44.2
            & \underline{48.1}
            & \underline{45.4}
            & \underline{46.7} \\

        Contrastive MM-RAG
            & 38.8
            & 36.2
            & 37.6
            & \underline{39.4}
            & 41.7
            & 39.6
            & 47.3
            & 45.8
            & 41.4
            & 40.3
            & 40.8 \\

        Native MM-RAG
            & 34.3
            & 29.6
            & 37.4
            & 34.6
            & 44.5
            & 44.3
            & 49.6
            & \underline{51.2}
            & 41.5
            & 39.9
            & 40.7 \\

        \addlinespace[0.5mm]

        M3-Agent
            & 19.5
            & 18.4
            & 17.2
            & 17.8
            & 16.6
            & 17.8
            & 20.6
            & 21.4
            & 18.5
            & 18.9
            & 18.7 \\

        M2A
            & 15.4
            & 14.2
            & 16.7
            & 18.6
            & 14.5
            & 17.6
            & 18.3
            & 19.6
            & 16.2
            & 17.5
            & 16.9 \\

        \midrule

        \textbf{PMMC (Ours)}
            & \underline{44.3}
            & \textbf{48.6}
            & \textbf{42.5}
            & \textbf{39.7}
            & \underline{56.2}
            & \textbf{57.8}
            & \textbf{58.6}
            & \textbf{51.4}
            & \textbf{50.4}
            & \textbf{49.4}
            & \textbf{49.9} \\

        \bottomrule
    \end{tabular}
    }

    \caption{
    Harmonized Judge scores (HJ, $0$--$100$; higher is better) on the
    primary generative evidence-recovery splits of MEMLENS
    (ML, $N=699$) and Mem-Gallery (MG, $N=1{,}446$) across four answer
    backbones. ML Avg. and MG Avg. are macro-averages over the four
    backbones, and Overall is the macro-average over all eight
    backbone--benchmark settings. Best results in each column are in
    \textbf{bold}, and second-best results are
    \underline{underlined}.
    }
    \label{tab:main_results}
\end{table*}

\subsection{Problem Formulation}
\label{sec:method_problem}

Let the multimodal interaction history be
$\mathcal{H}=\{m_1,\ldots,m_N\}$, where
\begin{equation}
\begin{aligned}
    m_i &=
    (x_i,\mathcal{I}_i,c_i,s_i,\ell_i,b_i,\mu_i),\\
    \mathcal{H}_{\leq t}
    &=\{m_i\in\mathcal{H}\mid b_i\leq t\}.
\end{aligned}
\label{eq:visible_history}
\end{equation}
Here, $x_i$ is conversational text, $\mathcal{I}_i$ denotes associated
source images, $c_i$ is an optional caption, $s_i$ and $\ell_i$ denote
session and temporal order, $b_i$ is the boundary at which the memory
becomes available, and $\mu_i$ stores metadata. Captions and embeddings
serve only as auxiliary indices; original utterances and images remain
the canonical evidence.

At consolidation boundary $t$, every PMMC component is restricted to
$\mathcal{H}_{\leq t}$ within the same conversation and dataset
snapshot. Benchmark queries, reference answers, annotated clues, and
evaluator-side evidence are excluded. PMMC constructs a Question Bank
$\mathcal{B}_t$ containing prospective questions and verified access
programs, rather than cached responses.

\subsection{Prospective Question and Program Compilation}
\label{sec:prospective_compilation}

For each newly consolidated anchor, PMMC constructs a bounded
compilation unit $U_t$ from the anchor and earlier supporting memories.
Support is selected from complementary relations, including temporal
proximity, lexical or entity overlap, shared images, and profile or
preference links, subject to text and image budgets. A deterministic
analyzer further identifies evidence opportunities $\Omega_t$, such as
visual attributes, image--text bindings, temporal updates, conflicts,
and multi-evidence relations. Only temporary aliases for visible text
spans and images are exposed to the model.

Conditioned on $(U_t,\Omega_t)$, the Questioner generates a bounded set
of canonical prospective questions, paraphrases, and source-support
claims:
\begin{equation}
    \widetilde{\mathcal{Q}}_t
    =
    Q_{\theta}(U_t,\Omega_t).
    \label{eq:questioner}
\end{equation}
The compiler validates grounding and cutoff consistency, removes
duplicates and previously observed questions, and derives each
candidate's question type, reasoning mode, required modalities, and
minimum evidence multiplicity. It also constructs a provisional answer
$\widehat a_j$ and source support
$\widehat{\mathcal{E}}_j$ exclusively from the cited visible memories.
These fields are compiler-private and are never obtained from benchmark
annotations.

For candidate $\widehat q_j$, the compiler constructs a requirement
contract $R_j$ specifying the modalities, capabilities, and evidence
multiplicity required for answering it. The Planner observes
$(\widehat q_j,R_j)$ and typed revision directives, but not
$\widehat a_j$, $\widehat{\mathcal{E}}_j$, source identifiers, or
benchmark labels. At refinement round $r$, it selects a constrained
strategy that is deterministically lowered into a typed program:
\begin{equation}
\begin{aligned}
    z_j^{(r)}
      &=P_{\phi}
        (\widehat q_j,R_j,\Delta_j^{(r-1)}),\\
    \pi_j^{(r)}
      &=\operatorname{Lower}(z_j^{(r)},R_j)
        =[o_{j,1}^{(r)},\ldots,o_{j,L_j}^{(r)}].
\end{aligned}
\label{eq:program_compilation}
\end{equation}
Programs draw from a frozen kernel of lexical and dense text retrieval,
cross-modal image retrieval, image--text and local-context expansion,
temporal expansion, raw-image materialization, and evidence-control
operators. The lowerer enforces bounded length, typed acyclic dataflow,
registered operators, and valid prospective lineage; arbitrary code,
SQL, paths, and unrestricted tool calls are disallowed.

\subsection{Execution-Grounded Verification}
\label{sec:program_verification}

PMMC verifies each initial or revised program using the same operator
kernel employed online:
\begin{equation}
    (T_j^{(r)},E_j^{(r)})
    =
    \operatorname{Exec}
    \left(
        \pi_j^{(r)},
        \mathcal{H}_{\leq t};
        \widehat q_j
    \right),
    \label{eq:program_execution}
\end{equation}
where $T_j^{(r)}$ is the execution trace and $E_j^{(r)}$ is the
retrieved typed evidence. A program is structurally ready only if it
executes successfully, returns non-empty evidence, reaches all required
modalities and capabilities, and satisfies the required source
multiplicity:
\begin{equation}
\begin{aligned}
    \operatorname{Ready}(E,T;R)
    ={}&
    \operatorname{ExecOK}(T)
    \land |E|>0\\
    &\land\,
    \mathcal{M}^{\mathrm{req}}(R)\subseteq\mathcal{M}(E)\\
    &\land\,
    \mathcal{K}^{\mathrm{req}}(R)\subseteq\mathcal{K}(T)\\
    &\land\,N(E)\geq g(R).
\end{aligned}
\label{eq:execution_ready}
\end{equation}
This execution-guided check prevents a plausible textual plan from
being accepted when its actual evidence path is invalid
~\cite{wang2018robust}.

The Doubter evaluates the prospective question, provisional answer,
private source support, program, trace, and retrieved evidence. It
returns \emph{pass}, \emph{revise}, or \emph{reject}. Revisable
failures are translated into typed directives, such as adding visual
retrieval, expanding temporal context, materializing the original
image, or increasing bounded retrieval depth. The Planner receives
these directives without access to the private answer or source
identifiers, and the revised program is executed again for at most
$r_{\max}$ transitions. Thus, unlike unconstrained self-refinement
~\cite{madaan2023self,shinn2023reflexion}, every PMMC revision is tied
to an observed execution failure and a non-overridable readiness check.

The runtime Question Bank retains only the canonical question,
paraphrases, safe type and modality metadata, requirement contract,
earliest passing program, and version information. Provisional answers,
private source support, Planner rationales, and Doubter feedback are
removed. Consequently, the Question Bank specifies \emph{how} memory
should be accessed but cannot directly provide an answer.

\begin{table*}[t]
    \centering
    {%
    \small
    \setlength{\tabcolsep}{5.0pt}
    \renewcommand{\arraystretch}{1.08}

    \begin{tabular}{lcccccccc}
        \toprule
        Variant
        & \multicolumn{2}{c}{HJ $\uparrow$}
        & F1 $\uparrow$
        & Ev.R $\uparrow$
        & Img.H $\uparrow$
        & Prog. $\uparrow$
        & QTok $\downarrow$
        & WTok $\downarrow$ \\
        \cmidrule(lr){2-3}
        & ML & MG & & & & & & \\
        \midrule

        \textbf{PMMC (full)}
            & \textbf{56.2}
            & \textbf{57.8}
            & \textbf{49.3}
            & \textbf{73.0}
            & \textbf{69.4}
            & \textbf{92.1}
            & 3.2
            & 31.8 \\

        \quad w/o Doubter
            & 53.4
            & 54.6
            & 46.5
            & 67.8
            & 64.8
            & 82.7
            & 3.8
            & \textbf{18.7} \\

        \quad w/o Dynamic Planner
            & 51.8
            & 52.6
            & 44.8
            & 64.7
            & 58.8
            & 76.9
            & 4.2
            & 25.9 \\

        \quad w/o Raw-Image Access
            & 53.9
            & 52.9
            & 46.2
            & 68.6
            & 0.0
            & 84.6
            & \textbf{2.7}
            & 29.4 \\

        \bottomrule
    \end{tabular}
    }

    \caption{
    Ablation results with Qwen3.5-9B on MEMLENS (ML) and Mem-Gallery
    (MG). HJ is reported separately for the two benchmarks, whereas
    F1, prompt evidence recall (Ev.R), raw-image evidence hit (Img.H),
    and program success (Prog.) are macro-averaged across them.
    Img.H is evaluated only on queries requiring visual evidence.
    QTok denotes average query-time model tokens per query and WTok
    denotes average write-time model tokens per compilation unit, both
    in thousands. All quality and success metrics are on a $0$--$100$
    scale. Higher is better except for QTok and WTok.
    }
    \label{tab:ablation}
\end{table*}

\begin{figure}[t]
    \centering
    \includegraphics[width=0.98\linewidth]{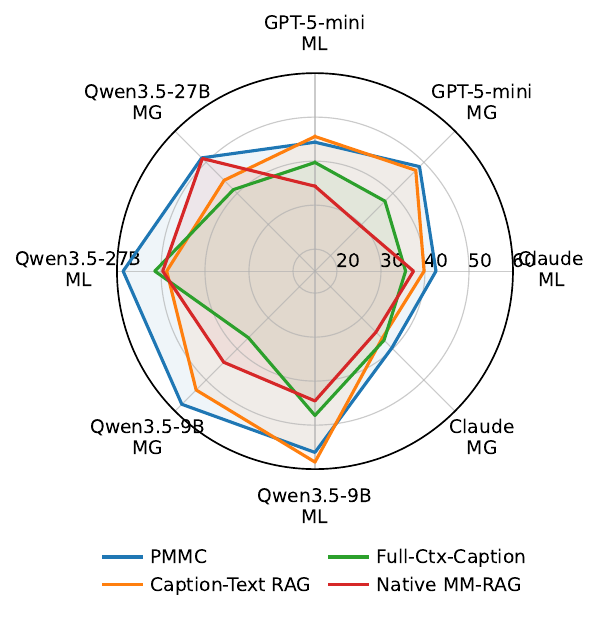}
    \caption{
    Radar-chart view of the main results using representative methods:
    PMMC, Caption-Text RAG, Full-Ctx-Caption, and Native MM-RAG.
    Each axis corresponds to one backbone--benchmark setting from
    Table~\ref{tab:main_results}; larger values indicate better
    Harmonized Judge performance. PMMC shows the strongest overall
    profile across the eight settings, with especially consistent
    gains on Mem-Gallery.
    }
    \label{fig:main_results_radar}
\end{figure}

\subsection{Online Routing and Answering}
\label{sec:online_execution}

For a real query $q$ with optional images $\mathcal{I}_q$ at boundary
$t_q$, PMMC first filters Question Bank entries by conversation,
snapshot, and cutoff compatibility. It then ranks candidates through
lexical, dense-text, and, when query images are present, multimodal
matching. The rankings are combined using weighted reciprocal-rank
fusion~\cite{cormack2009reciprocal}, and a compiled route is accepted
only when both its confidence and top-two margin pass fixed thresholds.

The selected program is validated and re-executed over
$\mathcal{H}_{\leq t_q}$. The incoming query determines which program
is selected but does not trigger replanning. Importantly, the program
freezes an access strategy and its retrieval representations, not a
fixed set of evidence identifiers. It can therefore retrieve valid
memories that became visible after its compilation.

A gold-free readiness check applies Eq.~\ref{eq:execution_ready} without
the provisional answer or private source support. If the check passes,
the retrieved references are materialized as original dialogue text,
captions, conversational context, and source images. The answer LVLM
receives the actual query, its query images, and only this
program-emitted evidence:
\begin{equation}
    A(q)=
    \begin{cases}
      \begin{aligned}
        &G\!\left(
          q,\mathcal{I}_q,\right.\\
        &\qquad\left.
          \operatorname{Materialize}(E_q)
        \right),\\
        &\text{if routing and readiness pass},
      \end{aligned}
      \\[2mm]
      \begin{aligned}
        &G_{\mathrm{MMRAG}}\!\left(
          q,\mathcal{I}_q,\right.\\
        &\qquad\left.
          \mathcal{H}_{\leq t_q}
        \right),\\
        &\text{otherwise}.
      \end{aligned}
    \end{cases}
    \label{eq:final_answer}
\end{equation}
The provisional answer $\widehat a_j$ is never included in the answer
prompt. Missing or ambiguous routes, invalid programs, execution
failures, and insufficient evidence all trigger the Native
Multimodal-RAG fallback.

PMMC therefore removes Questioner, Planner, Doubter, and iterative
access planning from the online critical path. For a compiled state
serving $n$ future queries, we report the amortized cost
\begin{equation}
    \overline C_{\mathrm{PMMC}}(n)
    =
    \frac{C_{\mathrm{compile}}}{n}
    +
    C_{\mathrm{online}},
    \label{eq:amortized_cost}
\end{equation}
where $C_{\mathrm{online}}$ includes routing, program execution,
fallback when invoked, and final answer generation.

\section{Experiment}

\subsection{Datasets and Evaluation}
\label{sec:datasets_and_evaluation}

We evaluate PMMC on two benchmarks for multimodal long-term
conversational memory. \textsc{MemLens} contains 789 questions covering
information extraction, multi-session and temporal reasoning, knowledge
updates, and answer refusal over interleaved textual and visual
histories~\cite{ren2026memlens}. We use its 32K setting and exclude the
90 answer-refusal instances, yielding 699 answerable queries.
\textsc{Mem-Gallery} contains 1,711 questions across 20 extended
multimodal histories and evaluates factual retrieval, visual search and
reasoning, test-time learning, temporal and multi-entity reasoning,
knowledge revision, conflict detection, and answer refusal
~\cite{bei2026mem}. Excluding the 81 conflict-detection and 184
answer-refusal instances leaves 1,446 queries for our primary
generative evidence-recovery evaluation. All methods within a reported
comparison use identical query identifiers and may access only memories
preceding the query-specific cutoff; reference answers and annotated
evidence are available only to the offline evaluator.

We compare against seven baselines: two long-context variants using
either captions or original images; Caption-Text RAG, Contrastive
Multimodal RAG, and Native Multimodal RAG; and benchmark adaptations of
M3-Agent~\cite{long2025seeing} and M2A~\cite{feng2026m2a}. We report a
method-blind Harmonized Judge score in $\{0,0.5,1\}$ together with
normalized exact match, token-level F1, and BLEU-1
~\cite{rajpurkar2016squad,papineni2002bleu}. Retrieval quality is
measured using evidence Recall/Hit@$K$, visual evidence hit, and
multi-evidence completeness. PMMC is further evaluated by Question-Bank
Coverage@$K$, program success, fallback rate, query-time tokens and
latency, write-time overhead, and amortized cost.

\begin{figure*}[t]
    \centering

    \begin{tabular}{@{}p{0.485\textwidth}@{\hspace{0.02\textwidth}}p{0.485\textwidth}@{}}
        \centering
        \includegraphics[width=\linewidth]{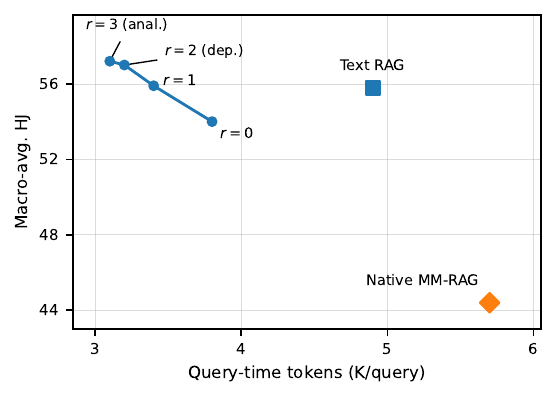}
        &
        \centering
        \includegraphics[width=\linewidth]{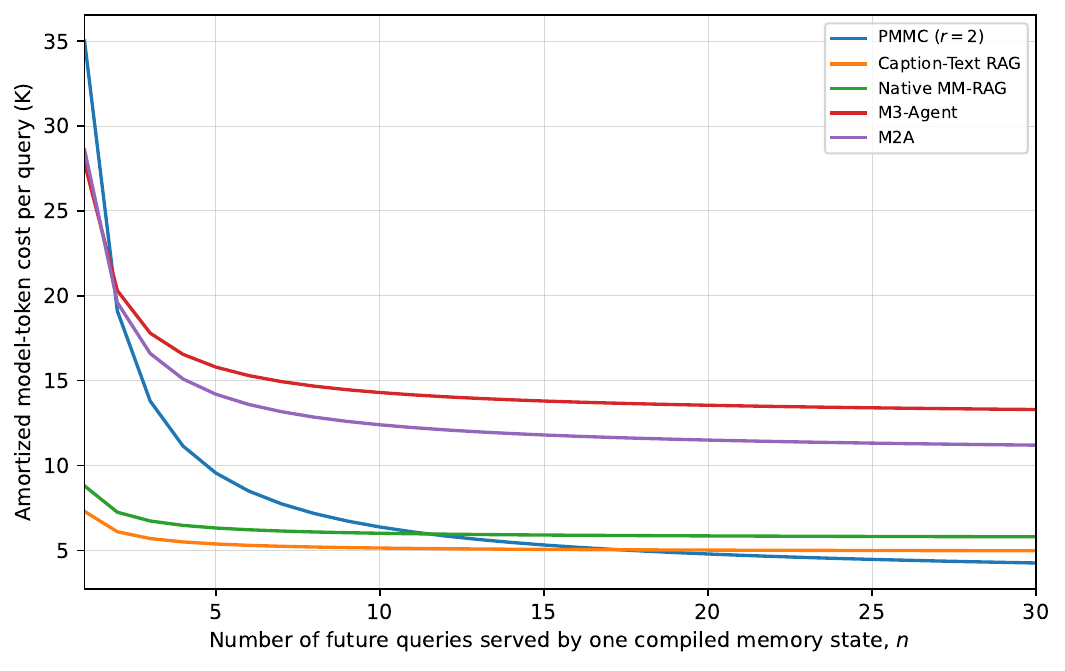}
        \tabularnewline
        \centering
        {\small (a) Accuracy--query-time cost trade-off.}
        &
        \centering
        {\small (b) Amortized write-plus-query cost.}
        \tabularnewline
    \end{tabular}

    \caption{
    Efficiency analysis under Qwen3.5-9B.
    In (a), HJ is macro-averaged over MEMLENS and Mem-Gallery, and
    the connected points show the effect of increasing PMMC's
    refinement limit. The $r=2$ point is deployed, whereas $r=3$ is
    retained only for analysis because of its higher write-time cost.
    Only the competitive fixed-retrieval baselines are shown in this
    panel; M3-Agent and M2A fall outside the displayed HJ range.
    Panel (b) reports
    $C_{\mathrm{write}}/n+C_{\mathrm{query}}$ for PMMC ($r=2$),
    Caption-Text RAG, Native MM-RAG, M3-Agent, and M2A, where $n$ is
    the number of future queries served by one compiled memory state.
    Costs are measured in thousands of model tokens.
    }
    \label{fig:self_feedback_efficiency}
\end{figure*}

\subsection{Implementation Details}
\label{sec:implementation_details}

We evaluate GPT-5-mini and Claude Haiku 4.5 as closed-source backbones,
and Qwen3.5-9B and Qwen3.5-27B as open-weight backbones. Within each
condition, PMMC uses the same backbone for the Questioner, Planner,
Doubter, and answer generator, with separate role prompts and structured
output schemas. GPT-5-mini serves as the fixed method-blind judge for
all methods and backbones. Caption-Text RAG uses
Qwen3-Embedding-8B, Contrastive Multimodal RAG uses
SigLIP2-So400M, and Native Multimodal RAG uses
Qwen3-VL-Embedding-8B~\cite{zhang2025qwen3,
tschannen2025siglip,li2026qwen3}. All embeddings are
$\ell_2$-normalized and compared using exact cosine similarity.

All final-answer calls use a 32,768-token context budget with at most
1,024 output tokens. The RAG baselines construct session-bounded
windows of at most 512 text tokens and four images, retain the top 50
results for retrieval evaluation, and provide the top five windows to
the answer model. PMMC uses a prospective-question budget of $b=8$, at
most two Planner--Doubter refinement transitions, and programs of no
more than eight operators. Question-Bank routes are fused using
reciprocal-rank fusion with $\kappa=60$ and are accepted at confidence
$\alpha_{\mathrm{route}}=0.60$ with a top-two margin of
$\delta=0.05$; otherwise PMMC falls back to Native Multimodal RAG.
Temperature is set to zero where supported. Open-weight models are
served in bfloat16 with vLLM~\cite{kwon2023efficient} on NVIDIA H100
80\,GB GPUs.

\subsection{Empirical Results}
\label{sec:empirical_results}

As shown in Table~\ref{tab:main_results}, PMMC achieves the highest
score in six of the eight backbone--benchmark settings and ranks
second in the remaining two. Averaged across the four backbones, it
reaches 50.4 on MEMLENS and 49.4 on Mem-Gallery, outperforming the
strongest baseline on average, Caption-Text RAG, by 2.3 and 4.0 points,
respectively; its overall average is 49.9, compared with 46.7 for the
strongest baseline. PMMC ranks first on Mem-Gallery with every
backbone, while on MEMLENS it leads with Claude Haiku 4.5 and
Qwen3.5-27B and remains second with GPT-5-mini and Qwen3.5-9B,
indicating that its gains are not tied to a single model family.

Figure~\ref{fig:main_results_radar} visualizes these trends using
representative baselines from the main comparison families. The radar
chart shows that PMMC encloses the broadest overall profile across the
eight backbone--benchmark settings, reflecting its consistently strong
performance rather than gains concentrated in a single condition.
Compared with Caption-Text RAG, PMMC is slightly weaker on
GPT-5-mini/MEMLENS and Qwen3.5-9B/MEMLENS, but clearly stronger on all
four Mem-Gallery settings and substantially stronger on
Qwen3.5-27B/MEMLENS. Full-Ctx-Caption remains competitive in a few
settings, especially with Qwen3.5-27B on MEMLENS, but is less stable
overall, while Native MM-RAG is more competitive on some
Mem-Gallery conditions than on MEMLENS. Full-Ctx-Caption also
consistently outperforms Full-Ctx-Raw in Table~\ref{tab:main_results},
suggesting that directly supplying historical images does not by
itself ensure effective use of multimodal evidence under a fixed
context budget. Overall, the results support prospective,
question-conditioned memory compilation as a more robust alternative to
committing to a single long-context or fixed-retrieval strategy.

\section{Analysis}

\subsection{Ablation Study}
\label{sec:ablation_study}

Table~\ref{tab:ablation} shows that all three components contribute to
PMMC. Removing the Doubter reduces the HJ score by 2.8 points on
MEMLENS and 3.2 points on Mem-Gallery, while decreasing program success
from 92.1 to 82.7; although this saves approximately 41\% of write-time
tokens, the less reliable programs require more online computation.
Replacing the question-conditioned Planner with a fixed retrieval plan
causes the largest overall degradation, reducing evidence recall by
8.3 points, raw-image hit by 10.6 points, and program success by
15.2 points. This result supports dynamically compiling different
access strategies for different prospective questions rather than
applying one retrieval policy universally. Finally, removing raw-image
access lowers HJ by 2.3 points on MEMLENS but by 4.9 points on the more
visually intensive Mem-Gallery, despite reducing query-time tokens by
15.6\%. Thus, captions and textual indices cannot fully replace access
to original visual evidence. Overall, PMMC obtains the strongest answer
quality and execution reliability at the intended cost of additional
memory-consolidation computation.

\subsection{Effect of Self-Feedback Rounds}
\label{sec:self_feedback_rounds}

\begin{table}[t]
    \centering
    {%
    \small
    \setlength{\tabcolsep}{3.2pt}
    \renewcommand{\arraystretch}{1.06}

    \begin{tabular}{lccccc}
        \toprule
        Setting
        & HJ $\uparrow$
        & Prog. $\uparrow$
        & FB $\downarrow$
        & QTok $\downarrow$
        & WTok $\downarrow$ \\
        \midrule

        $r=0$ (no refinement)
            & 54.0
            & 82.7
            & 22.4
            & 3.8
            & \textbf{18.7} \\

        $r=1$
            & 55.9
            & 89.1
            & 16.7
            & 3.4
            & 25.4 \\

        \textbf{$r=2$ (deployed)}
            & 57.0
            & 92.1
            & 12.6
            & 3.2
            & 31.8 \\

        $r=3$ (analysis only)
            & \textbf{57.2}
            & \textbf{92.7}
            & \textbf{12.2}
            & \textbf{3.1}
            & 38.5 \\

        \bottomrule
    \end{tabular}
    }

    \caption{
    Effect of the maximum number of Planner--Doubter refinement
    transitions under Qwen3.5-9B. HJ is macro-averaged over MEMLENS
    and Mem-Gallery. Prog. is the percentage of selected programs that
    execute successfully and pass the gold-free readiness check, and
    FB is the query-time fallback rate. QTok and WTok denote average
    query-time model tokens per query and write-time model tokens per
    compilation unit, respectively, in thousands. The bold row label
    marks the deployed setting; bold values are the best in each
    column.
    }
    \label{tab:self_feedback_rounds}
\end{table}

Table~\ref{tab:self_feedback_rounds} and
Figure~\ref{fig:self_feedback_efficiency} show that execution-grounded
feedback improves both program reliability and online efficiency.
Increasing the refinement limit from $r=0$ to $r=2$ raises the average
HJ score from 54.0 to 57.0 and program success from 82.7\% to 92.1\%,
while reducing fallback by 9.8 points and query-time tokens by 15.8\%.
A third transition provides only a further 0.2 HJ points and 0.6
program-success points, but increases write-time tokens by 21.1\%
relative to $r=2$; we therefore deploy $r=2$. This operating point
also lies above and to the left of the competitive fixed-retrieval
baselines in Figure~\ref{fig:self_feedback_efficiency}(a). Although
PMMC incurs the largest initial consolidation cost, its lower online
cost recovers the overhead after approximately two queries relative
to M3-Agent and M2A, twelve relative to Native MM-RAG, and eighteen
relative to Caption-Text RAG. These results support bounded
self-feedback as a favorable quality--cost trade-off for repeatedly
accessed long-term memory.

\subsection{Question Bank Coverage}
\label{sec:qbank_coverage}

\begin{table}[t]
    \centering
    {%
    \small
    \setlength{\tabcolsep}{4.8pt}
    \renewcommand{\arraystretch}{1.06}

    \begin{tabular}{lccc}
        \toprule
        Metric
        & MEMLENS
        & Mem-Gallery
        & Macro Avg. \\
        \midrule

        QBank-Cover@1
            & 52.1
            & 57.8
            & 55.0 \\

        QBank-Cover@3
            & 70.8
            & 76.2
            & 73.5 \\

        QBank-Cover@5
            & 79.6
            & 84.5
            & 82.1 \\

        \addlinespace[0.4mm]

        QBank-Match@1
            & 45.7
            & 50.9
            & 48.3 \\

        QBank-Match@3
            & 63.4
            & 69.7
            & 66.6 \\

        QBank-Match@5
            & 71.9
            & 78.4
            & 75.2 \\

        \addlinespace[0.4mm]

        Oracle-QBank HJ
            & 63.8
            & 66.9
            & 65.4 \\

        Fallback Rate
            & 14.6
            & 10.6
            & 12.6 \\

        \bottomrule
    \end{tabular}
    }

    \caption{
    Question-Bank coverage and routing diagnostics under Qwen3.5-9B.
    Cover@$K$ is the percentage of queries whose required source groups
    are fully covered by the union of the top-$K$ Question-Bank
    candidates. Match@$K$ indicates that at least one of the top-$K$
    candidates expresses an information need sufficient to answer the
    real query. Oracle HJ uses evaluator-side evidence labels only to
    select the most appropriate compiled candidate, after which the
    same frozen program and answer model are executed. FB denotes the
    deployed fallback rate. All values are percentages except Oracle
    HJ, which is reported on the $0$--$100$ scale.
    }
    \label{tab:qbank_coverage}
\end{table}

Table~\ref{tab:qbank_coverage} shows that the prospective Question Bank
covers a substantial, but not exhaustive, portion of future information
needs. Macro-averaged evidence coverage increases from 55.0\% at
$K=1$ to 82.1\% at $K=5$, while semantic matching rises from 48.3\% to
75.2\%. The 6.9-point gap between Cover@5 and Match@5 indicates that
some relevant evidence programs are already present in the bank but
are not ranked under a sufficiently compatible prospective question,
making routing---rather than question generation alone---an important
remaining bottleneck. Oracle candidate selection reaches an HJ score
of 65.4, 8.4 points above the deployed PMMC average of 57.0, revealing
additional headroom in candidate ranking, program selection, and
downstream answering. Nevertheless, only 12.6\% of queries invoke
fallback, suggesting that the Question Bank handles most queries
directly rather than succeeding through frequent recovery by
multimodal RAG.

\section{Conclusion}
\label{sec:conclusion}

We introduced \textbf{Prospective Multimodal Memory Compilation
(PMMC)}, a framework that reframes multimodal long-term memory as a
prospective compilation problem rather than relying exclusively on
query-time retrieval. During memory consolidation, PMMC generates
source-grounded prospective questions, compiles them into typed
multimodal memory programs, and verifies their evidence paths through
execution-grounded Planner--Doubter refinement. The resulting
Question Bank stores verified access strategies rather than predicted
answers; at query time, the selected program is re-executed over the
currently visible memory, with native multimodal RAG retained as a
fallback. Across MEMLENS and Mem-Gallery with four LVLM backbones, PMMC
ranks first in six of eight settings and improves the overall
macro-average Harmonized Judge score from 46.7 to 49.9 over the
strongest baseline. These results demonstrate that moving adaptive
memory-access planning to consolidation time can provide more robust
performance across heterogeneous multimodal information needs while
reducing dependence on repeated online planning. PMMC remains limited
by prospective-question coverage and additional write-time
computation, making its benefits most pronounced when consolidated
memories are accessed repeatedly. Future work may explore continual
Question-Bank maintenance, learned program policies, and extension to
richer modalities and interactive environments.




\bibliography{references}

\end{document}